\documentclass[12pt,english]{article}

\usepackage{mathptmx}
\usepackage{geometry}
\usepackage[fleqn]{amsmath}
\usepackage{pifont}
\usepackage{xcolor}
\definecolor{mygray}{gray}{0.6}
\usepackage{algorithm}
\usepackage{algorithmic}

\usepackage{array}
\usepackage{bbding}
\usepackage{multirow}
\usepackage{amssymb}
\usepackage{amsthm}
\usepackage{graphicx}
\usepackage{natbib}
\usepackage{lscape}
\usepackage[hyphens]{url}
\usepackage[colorlinks=true, citecolor=blue]{hyperref}
\usepackage{comment}
\usepackage{bm}
\usepackage[title]{appendix}
\usepackage{longtable}
\usepackage{mathtools}
\usepackage{chngcntr}
\usepackage{authblk}
\usepackage{babel}
\usepackage{dsfont}
\usepackage{fancyhdr}
\usepackage{cleveref}
\usepackage{abstract}

\usepackage{stackengine}
\usepackage{textcomp}
\usepackage{stfloats}
\usepackage{verbatim}
\usepackage{xcolor}
\usepackage{microtype}
\usepackage{graphicx}
\usepackage{subfigure}
\usepackage{booktabs} 
\usepackage{bbm}

\usepackage{amsfonts}
\usepackage{xurl}
\usepackage{stackengine}
\usepackage{tikz}
\usetikzlibrary{decorations.pathreplacing}
\usetikzlibrary{positioning,arrows.meta,quotes}
\usetikzlibrary{shapes,snakes}
\usetikzlibrary{bayesnet}
\tikzset{>=latex}
\tikzstyle{plate caption} = [caption, node distance=0, inner sep=0pt, below left=5pt and 0pt of #1.south]
\usepackage[normalem]{ulem}
\usepackage{multirow}\usetikzlibrary{positioning}

\usepackage{makecell}
\makeatletter
\allowdisplaybreaks
\date{}

\makeatletter
\def\blfootnote{\xdef\@thefnmark{}\@footnotetext}
\makeatother

\makeatletter
\def\titlepageext{
	\begin{center}	
	{\parindent0pt
		\rule{0.9\textwidth}{1pt}
		\begin{minipage}[t]{0.25\textwidth}
			\small {\it Keywords:}\\
			\keyword
		\end{minipage}%
		\hspace{3mm}
		\begin{minipage}[t]{0.6\textwidth}
			\small \abstract
		\end{minipage}%
		
		\rule{0.9\textwidth}{2pt}
	}
	\end{center}

	\blfootnote{* Corresponding author. E-mail address: \href{mailto:\corresemail}{\corresemail}.}
}
\makeatother

\usepackage[mathlines, pagewise]{lineno}
\usepackage{etoolbox} 

\newcommand*\linenomathpatchAMS[1]{%
	\expandafter\pretocmd\csname #1\endcsname {\linenomathAMS}{}{}%
	\expandafter\pretocmd\csname #1*\endcsname{\linenomathAMS}{}{}%
	\expandafter\apptocmd\csname end#1\endcsname {\endlinenomath}{}{}%
	\expandafter\apptocmd\csname end#1*\endcsname{\endlinenomath}{}{}%
}

\expandafter\ifx\linenomath\linenomathWithnumbers
\let\linenomathAMS\linenomathWithnumbers
\patchcmd\linenomathAMS{\advance\postdisplaypenalty\linenopenalty}{}{}{}
\else
\let\linenomathAMS\linenomathNonumbers
\fi

\linenomathpatchAMS{gather}
\linenomathpatchAMS{multline}
\linenomathpatchAMS{align}
\linenomathpatchAMS{alignat}
\linenomathpatchAMS{flalign}

\nolinenumbers

\usepackage{lineno}
\usepackage{graphicx}
\usepackage{subcaption}  
\usepackage{xcolor}
\usepackage{threeparttable}  

\title{NeuralMOVES: A lightweight and microscopic vehicle emission estimation model based on reverse engineering and surrogate learning}

\author[a,$\ast$]{Edgar Ramirez-Sanchez}
\author[a]{Catherine Tang}
\author[b]{Yaosheng Xu}
\author[a]{Nrithya Renganathan}
\author[a]{Vindula Jayawardana}
\author[a,$\ast$]{Zhengbing He}
\author[a]{Cathy Wu}
\affil[a]{Laboratory for Information \& Decision Systems, Massachusetts Institute of Technology, Cambridge MA, USA}
\affil[b]{Institute for Applied Computational Science, Harvard University, Cambridge MA, USA}

\def\corresemail{E. Ramirez-Sanchez (edgarrs@mit.edu), Z. He (he.zb@hotmail.com, hezb@mit.edu)}

\def\abstract{The transportation sector significantly contributes to greenhouse gas emissions, necessitating accurate emission models to guide mitigation strategies. Despite its field validation and certification, the industry-standard Motor Vehicle Emission Simulator (MOVES) faces challenges related to complexity in usage, high computational demands, and its unsuitability for microscopic real-time applications. 
To address these limitations, we present NeuralMOVES, a comprehensive suite of high-performance, lightweight surrogate models for vehicle CO$_2$ emissions. 
Developed based on reverse engineering and Neural Networks, NeuralMOVES achieves a remarkable 6.013\% Mean Average Percentage Error relative to MOVES across extensive tests spanning over two million scenarios with diverse trajectories and the factors regarding environments and vehicles. 
NeuralMOVES is only 2.4 MB, largely condensing the original MOVES and the reverse engineered MOVES into a compact representation, while maintaining high accuracy.
Therefore, NeuralMOVES significantly enhances accessibility while maintaining the accuracy of MOVES, simplifying CO$_2$ evaluation for transportation analyses and enabling real-time, microscopic applications across diverse scenarios without reliance on complex software or extensive computational resources. 
Moreover, this paper provides, for the first time, a framework for reverse engineering industrial-grade software tailored specifically to transportation scenarios, going beyond MOVES.
The surrogate models are available at \url{https://github.com/edgar-rs/neuralMOVES}.}

\def\keyword{Vehicle emission\\ MOVES\\ reverse engineering \\ surrogate modeling}

\begin{document}
\maketitle
\titlepageext

\newpage


\section{Introduction}

The transportation sector is one of the largest contributors to greenhouse gas (GHG) emissions worldwide, approximately accounting for one fourth of total CO$_2$ emissions \citep{EPA_GHG}. This significant contribution makes it a critical sector for climate change mitigation, as reducing emissions from transportation is essential for achieving global climate goals. The sector's transformation through electrification, automation, and intelligent infrastructure offers promising avenues for substantial emissions reductions \citep{Sciarretta_2020, EV_outlook, McKinsey_future_of_mobility}. However, the success of these innovations is critically dependent on the availability of suitable and accurate emission estimation models to guide the design and deployment of new technologies.

Motor Vehicle Emission Simulation (MOVES) \citep{USEPA_Motor_Vehicle_Emission_2022}, one of the most well-established emission estimation models, serves as the official and state-of-the-art emission estimation model in the U.S., provided, enforced, and maintained by the U.S. Environmental Protection Agency (EPA).
Despite its technical certification, MOVES' processing and software is tailored for two specific governmental uses: State Implementation Plans and Conformity Analyses \cite{EPA2021MOVES3}, which are for states to achieve and maintain air quality standards; and its use beyond trained practitioners and these specific analyses poses two main limitations.
First, a steep learning curve, computational demands, and complex inputs make it difficult for researchers and practitioners to use. 
In particular, MOVES has rigid input requirements, including a combination of toggle-based settings within its GUI and structured input files in specific formats. 
Second, MOVES is tailored for macroscopic analysis and is unsuitable for microscopic applications, such as control and optimization, which commonly require second-by-second emission calculations for individual actions and vehicles.
Unfortunately, many studies avoid using MOVES due to its technical complexities \citep{ravindra2006assessment,Kopfer2014ReducingGG,Zhao2018APB,wu2019tracking,Kou2020QuantifyingGG,Bai2022HybridRL,tsanakas2022generating}, despite the strong incentive to use a centralized, established, and certified model, such as MOVES, for accurate and comparable emission estimates.

To avoid this issue, alternative emission estimation models are used, with the choice of model depending on the application requirements \citep{8691686}. 
For high-level, aggregated analyses, users often choose macroscopic (link-based) models that are faster, more flexible, and easier to use. 
In contrast, microscopic (vehicle-based) applications, such as eco-driving \citep{mintsis2020dynamic,wu2019tracking} and trajectory planning \citep{10057033}, require real-time emission estimations for each vehicle action at a given time step. These models are crucial for emission analysis, leading to the emergence of various microscopic model alternatives \citep{Rakha_2011, hbefa, mkadziel2023vehicle}.
However, these alternatives account for different variables as inputs and their emission profiles vary significantly, making the estimation results incomparable. Moreover, optimization using different microscopic emission models may lead to vastly different outcomes \citep{Pre_VT_ecodirivng}. Therefore, using alternative microscopic models does not equate to utilizing the officially-recommended MOVES.

Given that alternative models cannot fully substitute MOVES, efforts have focused on preserving MOVES' processing capabilities while improving its usability. These efforts aim to bridge MOVES' emission modeling with applications that require faster, programmatic, and microscopic processing. In the literature, two MOVES variants are found, namely, MOVES-Matrix \citep{liu2016improved} and MOVEStar \citep{MOVESTAR}.
MOVES-Matrix is an emission modeling framework that precomputes MOVES outputs for all input combinations, storing them in a large lookup table. This allows for queries that retrieve MOVES-compliant emission estimates 200–800 times faster than running MOVES directly.
However, MOVES-Matrix inherits MOVES' complex input requirements, meaning users must still configure all MOVES parameters correctly. 
Additionally, the setup also requires high-performance computing infrastructure and massive storage (exceeding 100 GB per region), making it impractical for many real-time applications.
In contrast, MOVEStar uses a bottom-up approach to replicate the MOVES framework in a simplified manner. It provides a lightweight, microscopic alternative to MOVES while lacks accuracy measures and is limited to a reduced set of parameters, including only two vehicle types and fixed values for region, temperature, humidity, fuel type, and road grade.

\begin{figure}
    \centering
    \includegraphics[width=0.8\linewidth]{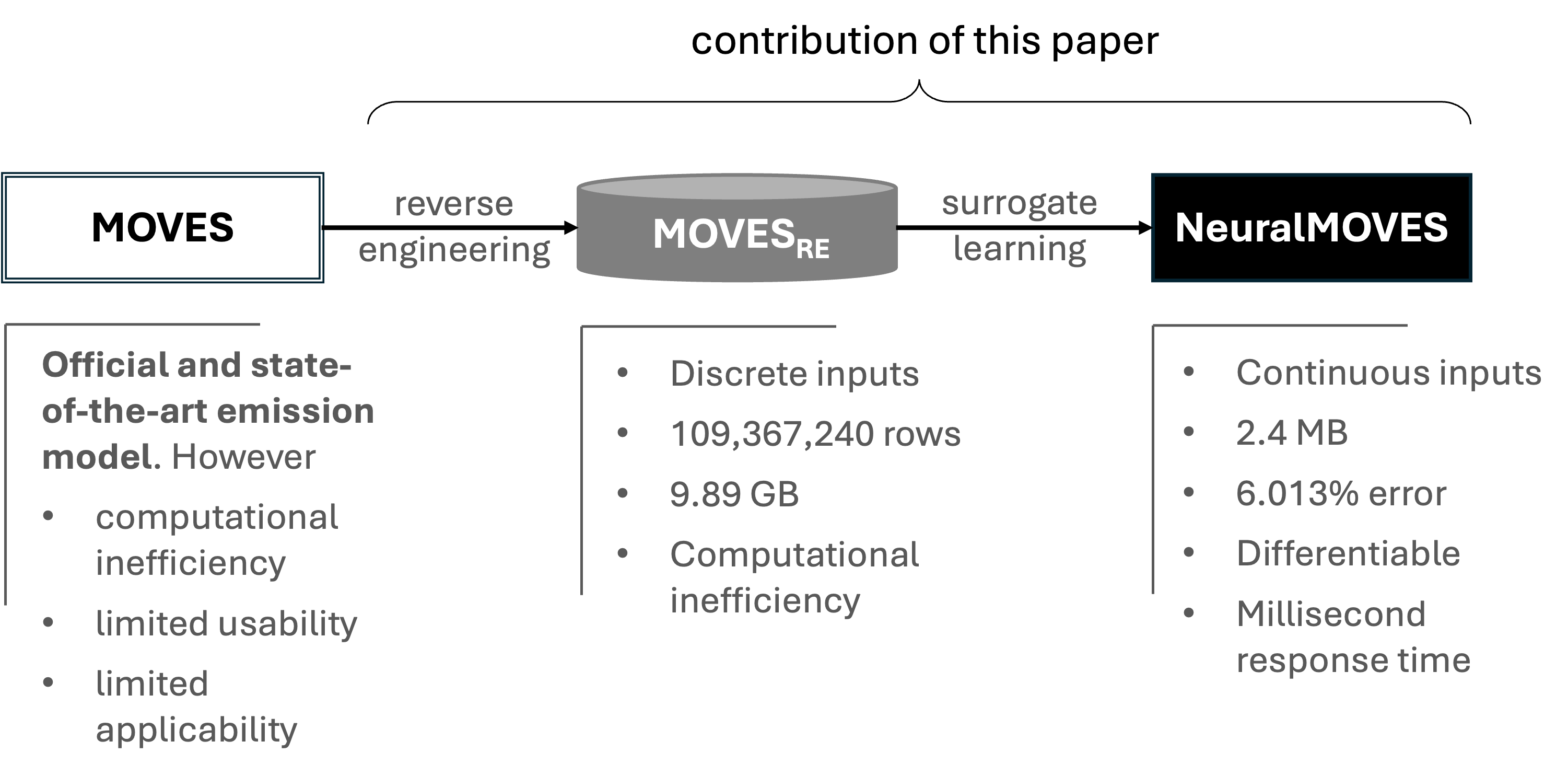}
    \caption{Contribution of the paper.}
    \label{fig:framework}
\end{figure}

To fulfill the significant practical demands and fill the gap left by the current MOVES variants, this paper introduces NeuralMOVES, a next-generation MOVES surrogate designed to provide microscopic models that are (1) sufficiently diverse to capture real-world conditions, with a comprehensive set of scenario parameters; (2) accurate enough to serve as a valid substitute for MOVES; and (3) lightweight and user-friendly, enabling real-time execution and accessibility for a wide range of users and use cases.
To this end, a reverse engineering-based method is first proposed to extract microscopic and instantaneous emissions from MOVES through careful scenario/input design, leading to a 9.89 GB dataset, named MOVES$_\text{RE}$ (Figure \ref{fig:framework}).
Surrogate learning, which is a widely used approach for approximating complex models or processes that are computationally expensive \citep{tao2019application, han2017weighted,westermann2019surrogate,van2019surrogate,CARNEVALE201647,AIhealth}, is then introduced to efficiently learn and reproduce gigabytes of data in MOVES$_\text{RE}$, reducing it to the lightweight NeuralMOVES with a concise set of Neural Network weight parameters. 
The result is a comprehensive family of microscopic CO$_2$ vehicle emission estimation models that cover a broad range of vehicle characteristics (vehicle type, fuel, age) and environmental conditions (road grade, temperature, humidity).
A comprehensive validation is conducted through extensive testing on over two million scenarios to compare NeuralMOVES with MOVES, assessing its accuracy and robustness. 
Additionally, we demonstrate the practical applicability of NeuralMOVES in control applications by optimizing eco-driving strategies at a signalized intersection.

NeuralMOVES is open-sourced at \url{https://github.com/edgar-rs/neuralMOVES}, and we hope the impact will be two-fold. 
First, by providing a faster, programmatic, and user-friendly alternative to MOVES, NeuralMOVES could simplify CO$_2$ emission evaluation in transportation-related analyses. 
Second, NeuralMOVES will bridge MOVES emission estimates with microscopic applications, such as vehicle-level control and optimization by offering more a lightweight, high-performance and microscopic version of MOVES.  

Moreover, this paper provides, for the first time, a framework for reverse engineering industrial-grade software specifically tailored to transportation-related emission scenarios. This framework provides an insight into how one could reverse engineer similar transportation software to adapt complex models into simpler, more efficient practical applications, provided that it is allowed under copyright law.

The rest of the paper is organized as follows. 
Section \ref{sec:Method} first introduces the methodology used to develop NeuralMOVES, including reverse engineering MOVES to obtain the MOVES$_\text{RE}$ dataset and then surrogate learning MOVES$_\text{RE}$.
Section \ref{sec:Validation} conducts a comprehensive validation analysis to compare the proposed NeuralMOVES with the original MOVES.
Section \ref{sec:UseCase} presents a case of dynamic eco-driving to show the capability of NeuralMOVES. 
Conclusions will be presented in Section \ref{sec:Conclusion}, along with future directions.

\section{Methodology}\label{sec:Method}

A two-step framework is proposed to develop NeuralMOVES: first, microscopic and instantaneous emission dataset, i.e., MOVES$_\text{RE}$, is extracted and collected through reverse-engineering MOVES; then, NeuralMOVES as a model is constructed by surrogate learning the reverse-engineered emission data (Figure \ref{fig:framework}).

\subsection{MOVES$_\text{RE}$: A reverse-engineered MOVES dataset}

The lowest (i.e., the most microscopic or granular) level at which MOVES can estimate emissions is a so-called \textit{driving cycle}, defined as a time series of speed in MOVES\footnote{
A related but distinct concept is a trajectory. 
A trajectory refers to the path of a vehicle in space and time, typically including information such as position (latitude/longitude or coordinates), speed, acceleration, and direction. Trajectory data represents the actual driving behavior of a vehicle on the road, capturing its dynamic movements.}. 
The speed could be the speed of a vehicle or the average speed of a group of vehicles. By using driving cycles, MOVES can model emissions at various scales, from individual vehicle trips to regional or national levels.
The goal of the reverse engineering is to extract instantaneous (second-by-second) emissions of a single vehicle from the driving cycle-based MOVES.

\subsubsection{Calculation of instantaneous emission}

We begin by defining a set of driving cycle tuples of the form \( \langle \boldsymbol\tau, \boldsymbol\tau' \rangle \), where \( \boldsymbol\tau \) and \( \boldsymbol\tau' \) represent the baseline and custom driving cycles, respectively.
The baseline driving cycle is defined as a time series with constant speed \( v \) and time intervals \( \Delta t \), ranging from 1 to \( n \), i.e.,
\begin{equation}
    \boldsymbol{\tau} = \{ v \}_{i=1}^{n}, \ v_i=v, \quad \forall i \in \{1, 2, \dots, n\}.
\end{equation}
%
The custom driving cycle \( \boldsymbol{\tau'} \) is formed by appending an additional time step \( n+1 \) to the baseline driving cycle, where the new speed \( v' \) at time step \( n+1 \) is determined based on a specific acceleration \( a \) of interest for one time step \( \Delta t \), i.e.,
\begin{equation}
    \boldsymbol{\tau'} = \{ v \}_{i=1}^{n}, \ v_i = v, \ v_{n+1}= v' = v + a \Delta t, \quad \forall i \in \{1, 2, \dots, n+1\}.
\end{equation}

Let \( E(\boldsymbol{\tau}, X) \) denote the total emissions of the driving cycle \( \boldsymbol{\tau} \), where \( X \) is the vector of other emission-related input factors, such as the environment and vehicle factors.
Since \( \boldsymbol\tau' \) is identical to \( \boldsymbol\tau \) for the first \( n \) time steps, differing only in the additional \( n+1 \) time step, where the speed transitions from \( v \) to \( v' \) with acceleration \( a \), we can compute the emission $e$ associated with applying an acceleration \( a \) at speed \( v \), as follows:
\begin{equation}\label{equ:emission}
   e(v, a, X)= E(\boldsymbol\tau', X) - E(\boldsymbol\tau, X).
\end{equation}

This method, which estimates instantaneous emissions by calculating the difference in emissions between two driving cycles, is particularly designed for MOVES, since MOVES is more suitable for calculating driving cycle-based emissions than directly calculating instantaneous emissions.
Moreover, MOVES computes different types of emissions for different operating stages, such as starting and running emissions. 
By adjusting the length of the baseline driving cycle \( \boldsymbol{\tau} \), i.e., \( n \), we can control which emission stage is the focus of the reverse engineering.
Here, we set $n=5$ and $\Delta t=1$ second, i.e., the resulting dataset will reflect the running emissions.
One may set $n=1$ to reflect the starting emissions.


Although we already have an $n$-second driving cycle, MOVES calculates emissions based on a standard 1-hour operating time. Therefore, to obtain the emissions for the $n$-second driving cycle, we need to follow a two-step process.

\begin{itemize}
    \item Run MOVES for 1-hour operation time. 
    We force MOVES to use a total operating time of 1 hour by setting the vehicle miles traveled to an artificially high value, which allows MOVES to assign the maximum processing duration, defaulting to $T = 1$ hour in MOVES' project mode, which is one of the multiple modes of MOVES. The emissions are calculated based on the vehicle's operating modes during this 1-hour period, and the operating mode distribution is influenced by the driving cycle's speed and acceleration patterns. Mathematically, the emissions for the 1-hour operating time $E_\text{1-hour}$ are computed as follows.
    \begin{equation}
        E_\text{1-hour}(\boldsymbol\tau, X) = T \sum_{j=1}^J p_j(\boldsymbol\tau) \cdot r_j(X),
    \end{equation}
    where $p_j(\boldsymbol{\tau})$ is a function that calculates the fraction of time spent in operating mode $j$ ($1\leq j\leq J$) from the driving cycle $\boldsymbol{\tau}$; and $r_j(X)$ is a function that specifies the emission rate for operating mode $j$ based on emission-related input factors required by MOVES.
    For $\boldsymbol{\tau}$, $J=1$ and $p_j=1$, since all the speeds in $\boldsymbol{\tau}$ are the same.
    For $\boldsymbol{\tau}'$, $J=2$, $p_1=(n-1)/n$ and $p_2=1/n$.

    \item Scale emissions to match the $n$-second cycle. Once we have the emissions for the 1-hour operation time, we can then scale them to the desired driving cycle length (e.g., $n$ or $n+1$ seconds). To scale the emissions, we use the fact that the emissions calculation in MOVES is linearly proportional to the operating time. Therefore, the emissions for the $n$-second cycle, i.e., $E$, can be estimated as follows.
    \begin{equation}
        E(\boldsymbol\tau, X) = \frac{E_\text{1-hour}(\boldsymbol\tau, X) \cdot n}{3600}.
    \end{equation}
\end{itemize}

\subsubsection{Dataset generation}\label{sec:DatasetGeneration}

There are many factors that influence vehicle emissions.
In addition to vehicle dynamics factors (i.e., speed ($v$) and acceleration ($a$)), other influential factors that are selected as the input include environment factors (i.e., road grade ($x_\text{grade}$), temperature ($x_\text{temp}$), and humidity ($x_\text{humid}$)), and vehicle factors (i.e., vehicle type ($x_\text{type}$), vehicle age ($x_\text{age}$), and fuel type ($x_\text{fuel}$)). 
Therefore, the input $X$ in Equation \ref{equ:emission}, which is used to reverse engineer MOVES, is concretized as
\begin{equation}\label{equ:X}
    X= \{x_\text{grade}, \ x_\text{temp}, \ x_\text{humid}, \ x_\text{type}, \ x_\text{age}, \ x_\text{fuel}\}.
\end{equation}

To construct a dataset, which is discrete in nature, we first discretize all above factors (Table~\ref{table:dimensions}) and then input each valid combination of the factor values to MOVES according to Equation \ref{equ:emission}.
Finally, we obtain the reverse-engineered dataset representing MOVES, named {MOVES}$_\text{RE}$ and written as
\begin{equation}\label{equ:MOVES_RE}
    \text{MOVES}_\text{RE}: (v, \ a, \ x_\text{grade}, \ x_\text{temp}, \ x_\text{humid}, \ x_\text{type}, \ x_\text{age}, \ x_\text{fuel}) \mapsto e.
\end{equation}
where all variables are discrete,and the total size of the dataset is determined as
\begin{equation}
\begin{aligned}
    |\text{MOVES}_\text{RE}| &= n_\text{speed,acc} \cdot n_\text{temp,humid} \cdot n_\text{grade} \cdot n_\text{age}  \cdot \overline{n}_\text{type}^\text{mc} \cdot n_\text{fuel} \\
    &+ n_\text{speed,acc} \cdot n_\text{temp,humid} \cdot n_\text{grade} \cdot  n_\text{age} \cdot n_\text{type}^\text{mc} \cdot n_\text{fuel}^\text{gas} \\
    &- n_\text{miss} = 109,367,240
\end{aligned}
\end{equation}
where $n_\text{speed,acc}$ is the number of the valid combinations of speed and acceleration;
$n_\text{temp,humid}$ is the number of the valid combinations of temperature and humidity; $n_\text{age}$ is the number of vehicle age values; 
$n_\text{type}^\text{mc}$ and $\overline{n}_\text{type}^\text{mc}$ are the numbers of the types of motorcycle and the total types of other vehicles, respectively;
$n_\text{fuel}$ and $n_\text{fuel}^\text{gas}$ are the numbers of the fuel types of fuel and the fuel types of gasoline;
$n_\text{miss}$ is the number of missing and corrupted values.
The specific values of the variables, which are determined based on the statistics in Table \ref{table:dimensions}, are listed as follows.
\begin{itemize}
    \item \(n_\text{speed,acc} = 4,791\). This is less than the product of the number of speed values and acceleration values (i.e., \(66 \times 76 = 5,016\)) due to invalid combinations. An invalid combination $v$, $a$, $v'$ happens the vehicle is subject to an negative acceleration value $a$ larger that its current speed $v$, since it would lead to a negative speed $v'$ . 

    \item \( n_{\text{temp,humid}} = 21 \). The temperature and humidity variables do not span the full combinatorial space, i.e., they are not multiplied such that every temperature value pairs with all 21 humidity values and vice versa.
    Instead, we select the combinations as follows.
    First, 10 combinations were carefully chosen to represent the annual average temperature and humidity levels of 10 U.S. cities \citep{jayawardana2024intersectionzoo}. Then, 11 additional points were strategically added to optimize coverage in the temperature-humidity space, using a minimization optimization process to effectively fill data gaps.
    Finally, we obtain only 21 unique temperature-humidity combinations\footnote{The temperature (°F) and relative humidity (\%) pairs finally selected for the dataset are 28.1996-80.2873, 30.0000-56.0000, 33.0000-32.0000, 37.0000-89.0000, 37.6286-67.6447, 43.0000-47.0000, 46.1619-80.2282, 53.6926-58.5524, 55.0000-33.0000, 55.6191-75.1602, 60.0000-68.0000, 65.0137-62.8505, 69.9007-82.7532, 70.0000-25.0000, 71.4719-46.6703, 72.0000-54.0000, 79.0944-62.6927, 82.0000-49.0000, 82.8811-26.8811, 87.0000-80.0000, and 89.0000-38.0000.}.

    \item $n_\text{type}^\text{mc}=1$ and $\overline{n}_\text{type}^\text{mc}=4$. The sum is 5, i.e., the total number of vehicle types in Table \ref{table:dimensions};
    
    \item $n_\text{fuel}=2$ and $n_\text{fuel}^\text{gas}=1$;

    \item $n_\text{age}=11$, including the latest model year available in MOVES: 2019; and the 10 years prior; 
  
    \item \( n_{\text{miss}} = 198,139 \). Throughout the extraction process, system crashes and failures occasionally required rerunning batches. While most missing data was recovered, a final inspection revealed that a subset of results contained errors due to an extraction-related technical issue. Specifically, for the group comprising vehicle age 2016, only 18 valid temperature-humidity combinations were successfully retrieved for motorcycles with gasoline. In total, 198,139 data points were identified as corrupted and removed from the dataset.     
\end{itemize}
It is worth noting that the discreteness and the small number of missing values in the inputs will not cause significant issues, as Section \ref{sec:Surrogate} introduces surrogate learning to connect the discrete inputs, effectively creating a continuous input space.

\begin{table}[ht]
\centering
\footnotesize
\setlength\tabcolsep{5.5pt} 
\caption{List of factors that are used to reverse engineer MOVES.\vspace{-2mm}}
\begin{tabular}{llllrrrrr} 
\hline
Type & Factors     & Notation       & Category/Unit        & Min    & Max    & Resolution & Number of Values \\ 
\hline
\addlinespace[1mm] 
\multirow{2}{*}{Dynamics}  & Speed & $v$ & m/s & 0 & 33 & 0.5 & \multirow{2}{*}{4,791} \\
& Acceleration & $a$ & m/s\(^2\) & -4.5 & 3 & 0.1 &  \\
\addlinespace[2mm] 
 & Road Grade   & $x_\text{grade}$  & \%  & -25 & 25  & 5   & 11     \\ 
Environment & Temperature & $x_\text{temp}$ & $^\circ$F & 28 & 90 & 5 $\sim$ 20 & \multirow{2}{*}{21} \\
& Relative Humidity & $x_\text{humid}$ & \% & 25 & 90 & 5 $\sim$ 20 & \\
\addlinespace[2mm] 
\multirow{8}{*}{Vehicle} & \multirow{2}{*}{Fuel} & \multirow{2}{*}{$x_\text{fuel}$} & Gasoline &  &  &  & \multirow{2}{*}{2} \\
& & & Diesel &  &  &  &  \\
\addlinespace[2mm] 
& \multirow{5}{*}{Type} & \multirow{5}{*}{$x_\text{type}$}  & Motorcycles & & & & \multirow{5}{*}{5} \\
& & & Passenger Cars & & & & \\
& & & Passenger Trucks & & & & \\
& & & Light Commercial Truck & & & & \\
& & & Transit Bus & & & & \\
\addlinespace[2mm] 
& Age   & $x_\text{age}$  & Year  & 2009 & 2019 & 1  & 11    \\ 
\addlinespace[1mm] 
\hline
\end{tabular}
\label{table:dimensions}
\end{table}

The overall extraction process involves setting up batches of runs, monitoring their execution, and post-processing the results. Each run requires generating all necessary input files, feeding them into MOVES, and executing the computations. MOVES processes emissions in `links,' which represent independent road segments. We leverage this to simulate multiple driving cycles per run. Once complete, the output data was stored in SQL databases, where further processing was required to filter and format the data, identify driving cycle pairs, scale emissions to the correct duration, and subtract baseline emissions to isolate instantaneous effects.

To optimize this workflow, we develop automated scripts to generate input files, interact with MOVES via the command line, and batch-process runs efficiently. By structuring runs strategically and utilizing MOVES' batch processing capabilities, we maximize throughput. However, despite these optimizations, the extraction process remained computationally expensive, running continuously for several weeks on four high-performance workstations, with occasional failures that required manual intervention.

This computational burden underscores the impracticality of relying on MOVES for microscopic emission modeling. It further motivates the development of a surrogate model that can replicate MOVES' output at a fraction of the computational cost, making high-resolution emission estimation more accessible and scalable.

\subsubsection{Dataset analysis}

\begin{figure}
    \centering
    \includegraphics[width=\linewidth]{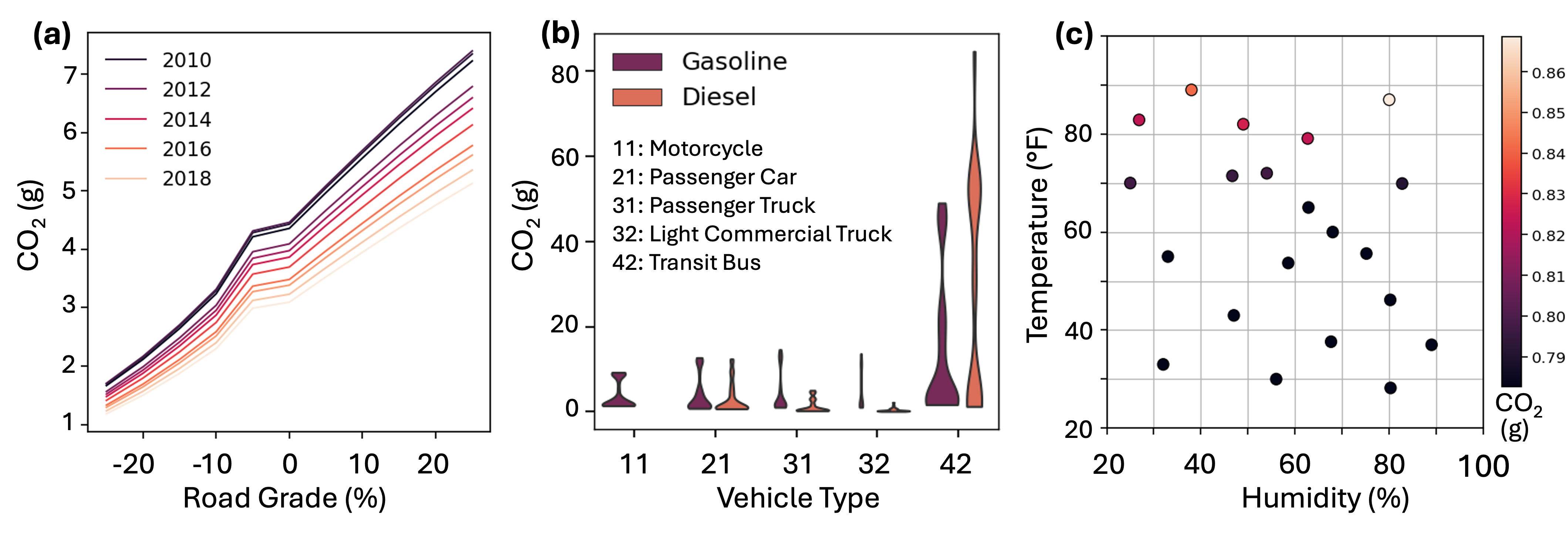}
    \caption{Impact of environment and vehicle factors on the instantaneous CO$_2$ emissions extracted by reverse-engineering MOVES.
    (a) Vehicle age and road grade;
    (b) Vehicle type and fuel type (Note: no diesel for motorcycle);
    (c) Temperature and humidity (Note: only 21 combinations are selected as we stated in Section \ref{sec:DatasetGeneration}).}
    \label{fig:emission}
\end{figure}

Figure \ref{fig:emission} presents the impact of environment and vehicle factors on the instantaneous CO$_2$ emissions extracted by reverse-engineering MOVES and the following observations are made.
\begin{itemize}
    \item Road grade emerges as one of the most significant factors, with steep grades leading to emissions increasing by up to four times, i.e., CO$_2$ emissions at the maximum grade are four times higher than those at the minimum grade (Figure \ref{fig:emission}(a)).
    Note that an emission surge is observed at a road grade of -5\%, the cause of which remains unclear despite a careful review of the code to ensure its correctness.
    
    \item Diesel-powered trucks emit significantly less than their gasoline-powered counterparts, while transit buses exhibit the highest emissions among all vehicle types (Figure \ref{fig:emission}(b)). 
    
    \item Existing alternative models typically ignore weather conditions and vehicle ages. However, our results demonstrate that temperature and humidity can still affect emissions by up to 10\% across the selected combinations (Figure \ref{fig:emission}(c)), while vehicle age can lead to up to a 30\% variation (Figure \ref{fig:emission}(a))
    These observations emphasize the importance of accounting for the diversity in vehicle types, environmental conditions, and weather when modeling emissions.
    
\end{itemize}

\begin{figure}
    \centering
    \includegraphics[width=\linewidth]{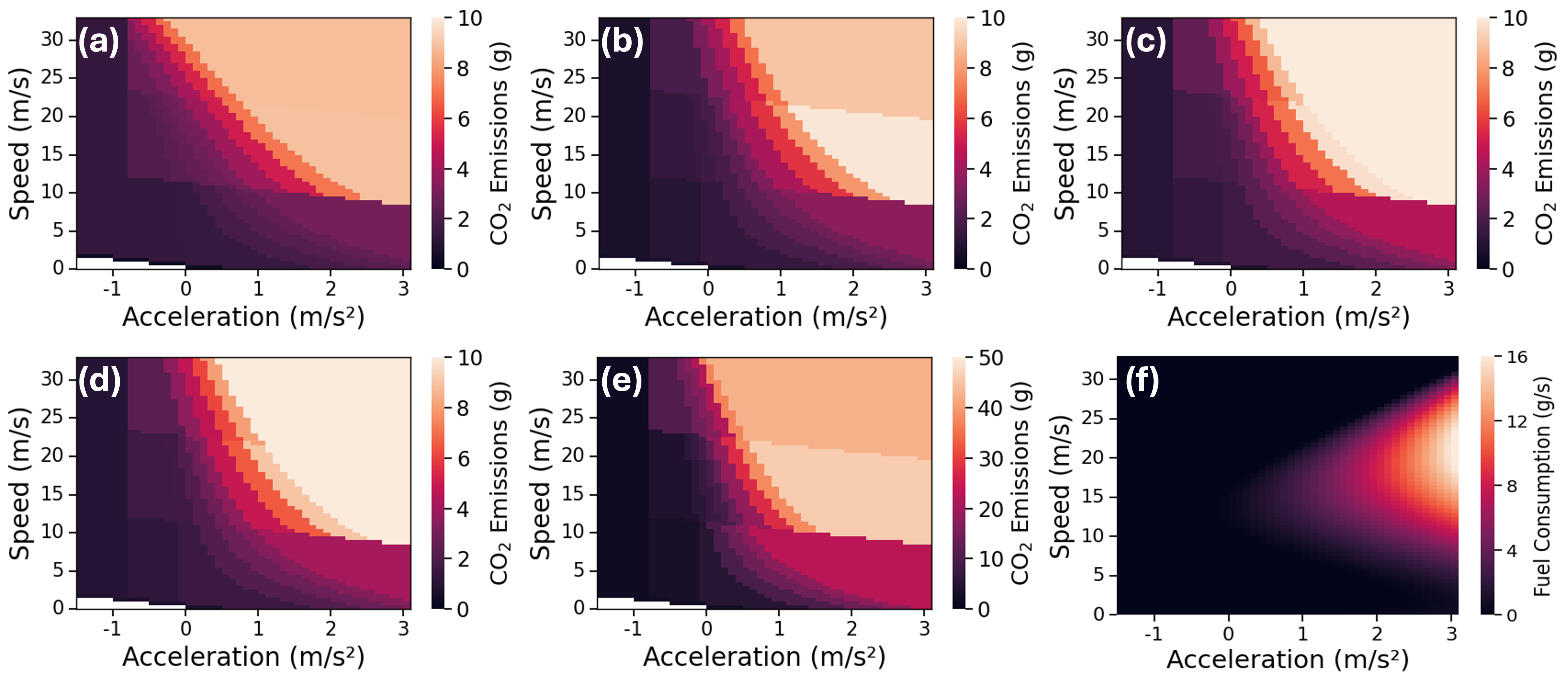}
    \caption{Impact of vehicle dynamics on the instantaneous CO$_2$ emissions extracted by reverse-engineering MOVES, characterized by vehicle types. (a) Motorcycles; (b) Passenger cars; (c) Passenger trucks; (d) Light commercial truck; (e) Transit bus; (f) VT-CPFM \citep{PARK2013317}.}
    \label{fig:speed_acceleration}
\end{figure}

Moreover, Figure \ref{fig:speed_acceleration} shows the impact of vehicle dynamics on the instantaneous CO$_2$ emissions extracted by reverse-engineering MOVES, characterized by vehicle types.
It confirms that emissions are sensitive to variations in speed and acceleration, with the specific sensitivities differing by vehicle types (Figure \ref{fig:speed_acceleration}(a)-(e)).
When comparing the results from the VT-CPFM model \citep{PARK2013317} (Figure \ref{fig:speed_acceleration}(f)), significant differences in the profiles are observed. These distinctions highlight that applications based on different datasets can yield divergent results. 
As such, researchers and practitioners should be particularly mindful of these discrepancies, as they can significantly impact both model predictions and real-world applications.

\subsubsection{Dataset features}

Different from MOVES, the reconstructed dataset has the following important features. First, it provides highly disaggregated data, capturing instantaneous emissions as a function of a specific speed and acceleration combinations for a single vehicle under given other static conditions (i.e., $X$ in Equations \ref{equ:emission} and \ref{equ:X}). In contrast, MOVES operates at a much more aggregated level. Second, this dataset is location-independent, enabling universal applicability. MOVES, by design, requires a U.S. location to infer variables such as temperature, humidity, fuel distribution, and vehicle fleet composition. By decoupling emissions data from location, the dataset allows users to directly input environmental parameters, facilitating emissions estimation for international contexts or customized scenarios, such as seasonal variations. This flexibility extends MOVES’ scope, offering enhanced control and granularity for both domestic and global applications, particularly for microscopic, real-time analyses.

The resulting instantaneous emission dataset contains over 100 million data points spanning all factors listed in Table \ref{table:dimensions}. It weighs 9.89 GB and is publicly available. While having instantaneous emissions is a significant advantage, and the dataset could be used as a lookup table, its format presents two important limitations when used directly.
First, the approximately 10 GB size can slow down applications, particularly when frequent data extraction and lookup are required in large-scale operations. 
Second, the discreteness of the input variables limits the widespread application of the dataset.
This necessitates interpolation or nearest-neighbor approaches to estimate unobserved points.
To address these issues, the following section will introduce surrogate modeling to significantly reduce the dataset size and convert it into a continuous function.

\subsection{NeuralMOVES: A learning-based surrogate of MOVES$_\text{RE}$}\label{sec:Surrogate}

Surrogate learning is a widely used approach for approximating complex models or processes that are computationally expensive. 
The fundamental concept involves training a surrogate model to approximate the output of a more detailed and computationally intensive model. 
By learning an accurate surrogate, practitioners can evaluate approximations of the target model more efficiently while preserving critical input-output relationships \citep{Cozad_2014}.

This section aims to replace the large, discrete dataset {MOVES}$_\text{RE}$ with a lightweight, continuous approximation using a Machine Learning (ML) model, enabling faster, real-time application.
The ML model, denoted by \( f_\theta \), is trained to learn the mapping from the inputs to the output by minimizing a loss function, which is the Mean Absolute Percentage Error (MAPE) here.
\begin{equation}
\text{MAPE}(\theta) = \frac{1}{N} \sum_{i=1}^{N} \left| \frac{e_i - f_\theta(x_i)}{e_i} \right| \times 100    
\end{equation}
where \( f_\theta(x_i) \) is the predicted output from the surrogate model and \( e_i \) is the output from {MOVES}$_\text{RE}$; 
$x_i$ is the input vector for the $i$-th training sample, i.e., $x_i = (v_i, a_i, x_{\text{grade},i}, x_{\text{temp},i}, x_{\text{humid},i}, x_{\text{type},i}, x_{\text{age},i}, x_{\text{fuel},i})$;
The model parameters \( \theta \) are optimized during training to minimize the loss function.
Note that the optimized output is additionally truncated with the idling emission value for that environment (denote by $e_\text{idling}$), which is the emission value associated with having the motor on without exerting any power. 
This truncation ensures that the minimum possible emission is the idling value, rather than zero or negative, which the model might predict when extrapolating the data.

For the surrogate model \( f_\theta \), we employ different model types and architectures, including polynomial regression with a third-order polynomial following the structure of alternative emission models \citep{hbefa, HBEFA_Software} as the baseline; and then decision trees and Neural Networks.
The following observations are made in the results presented in Table \ref{table:ablations}.
\begin{itemize}
    \item {Polynomial models} provide a simple representation but lack the capacity to capture the intricate dynamics of the data, with a high MAPE of 31.04\%.

    \item Decision Trees partition the input space into discrete regions, which can be suboptimal for continuous variables like temperature and humidity. While they achieve competitive accuracy with an MAPE of 6.17\%, the Mean Percentage Error (MPE) is more than double that of Neural Networks, indicating a more pronounced skew in the error distribution. Additionally, Decision Trees are non-differentiable, limiting their applicability in optimization and control tasks.

    \item {Neural Networks} emerge as the most effective architecture, capturing both the stepwise emissions pattern and continuous environmental variations. Through hyperparameter optimization, we identify an optimal architecture comprising two layers with five hidden neurons and a tanh activation function. This model achieves an MAPE of 6.01\%, closely approximating {MOVES}$_\text{RE}$ while remaining lightweight and highly generalizable.
    
\end{itemize}

\begin{table}[ht]
\centering
\footnotesize
\caption{Surrogate model architectures and ablations with end-to-end error statistics.}
\label{table:ablations}
\setlength\tabcolsep{8pt} 
\begin{tabular}{llrrrrr}
\toprule
Model Type            & Architecture                      & Training (ep)   & MAPE (\%)     & MPE (\%)      & MdPE (\%)     & StdPE        \\ 
\midrule
Polynomial            & 3rd order                         & -               & 31.04         & 8.9           & 5.22          & 50.16       \\
Decision Tree         & Depth 50                          & -               & 6.17          & 5.37          & 3.78          & \textbf{7.75} \\
\multirow{4}{*}{Neural Network} & 2 layers, dim 5                   & 11              & 149.73        & 124.43        & 65.64         & 165.17      \\
                                 & 3 layers, dim 64                  & 11              & 11.54         & 10.91         & 8.63          & 11.56       \\
                                 & 2 layers, dim 5                   & 300             & 7.85          & 6.04          & 4.81          & 9.44        \\
                                 & 2 layers, dim 5, scaling 0.97     & 300             & \textbf{6.01} & \textbf{2.46} & \textbf{1.22} & 8.90        \\
\bottomrule
\end{tabular}
\begin{tablenotes}\footnotesize
\item[1] \hspace{-2mm} MdPE: Median Percentage Error; StdPE: Standard Deviation of Percentage Errors
\end{tablenotes}  
\end{table}

Finally, a Neural Network with two layers, each consisting of 5 neurons (dim 5), and an initialization scaling factor of 0.97 is chosen as the surrogate architecture. 
Mathematically, the model can be written as follows:
\begin{equation}
    \text{NeuralMOVES}: \max\{e_\text{NN}, e_\text{idling}\} \mapsto e.
\end{equation}
where 
\begin{equation}
    e_\text{NN}=\text{NN}(v, \ a, \ x_\text{grade}, \ x_\text{temp}, \ x_\text{humid}, \ x_\text{type}, \ x_\text{age}, \ x_\text{fuel})
\end{equation}
where \( v \in [ \underline{v}, \overline{v} ] \),
\( a \in [ \underline{a}, \overline{a} ] \),
\( x_\text{grade} \in [ \underline{x}_\text{grade}, \overline{x}_\text{grade} ] \),
\( x_\text{temp} \in [ \underline{x}_\text{temp}, \overline{x}_\text{temp} ] \),
and \( x_\text{humid} \in [ \underline{x}_\text{humid}, \overline{x}_\text{humid} ] \).
The underline and overline represent the minimum and maximum values of a variable, respectively; and those values are listed in Table \ref{table:dimensions}.

Compared to {MOVES}$_\text{RE}$ in Equation \ref{equ:MOVES_RE}, {NeuralMOVES} has the following advantages.
First, {NeuralMOVES} updates its variables $v$, $a$, $x_\text{grade}$, $x_\text{temp}$, $x_\text{humid}$ to continuous ranges, enabling broader applications.
Second, a critical advantage of NeuralMOVES is its efficiency in data representation. {MOVES}$_\text{RE}$, comprising over 100 million data points and spanning about 10 gigabytes, is compressed into a Neural Network representation of just 2.4 MB. This corresponds to a compression of 4800$\times$ reduction in data size. Such an efficient representation not only reduces storage and computational requirements but also enables real-time emissions modeling and large-scale analysis.
In summary, by leveraging surrogate learning, we achieved accurate, compact, and microscopic representations of MOVES emissions data, enabling the integration of emissions modeling into optimization, control, and transportation analysis applications.

\section{Validation}\label{sec:Validation}

\begin{figure}
    \centering
    \includegraphics[width=1\textwidth]{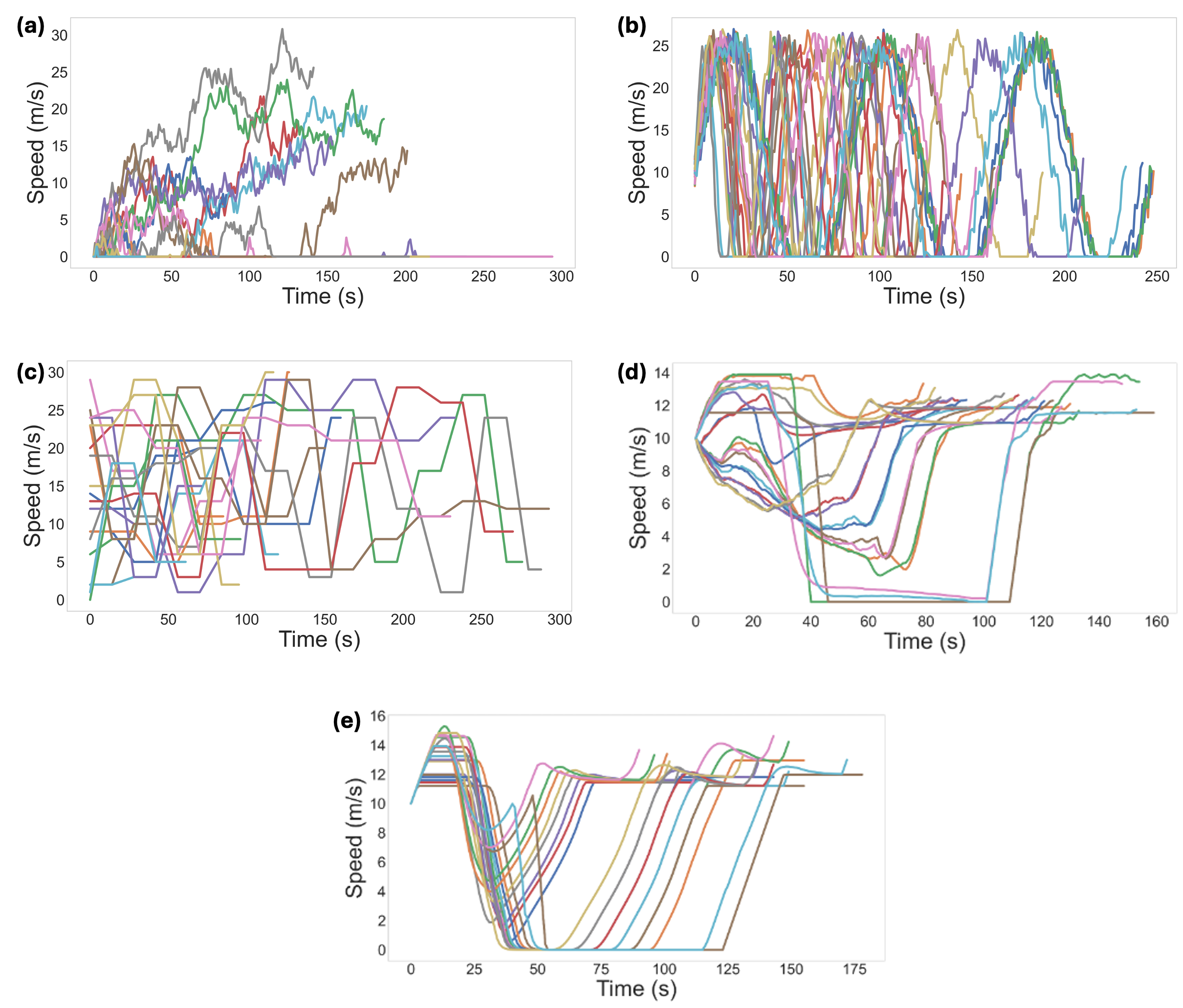}
    \caption{Synthetic driving cycles for validating NeuralMOVES. (a) Random speed; (b) Sinusoidal speed; (c) Piecewise speed; (d) Approaching intersection; (e) Eco-driving.}
    \label{fig:traj}
\end{figure}

To assess the accuracy of NeuralMOVES in representing MOVES, we conduct a direct comparison of NeuralMOVES emission outputs with MOVES outputs for an extensive test set.
The design of the validation set is based on representativeness ( i.e. including realistic and diverse tests), and robustness (i.e. having enough samples to properly draw conclusions). 
The set of evaluations are designed by varying two primary components: (1) environments and vehicles (2) driving dynamics; as the three types of factors detailed in Table \ref{table:dimensions}. 
The \textit{environments and vehicles} under which a vehicle operates are generated by systematically combining values for all factors regarding environments and vehicles in Table \ref{table:dimensions} within the minimum and maximum values.
This process renders a diverse set of scenarios that reflect a comprehensive evaluation of the model’s performance across diverse but realistic conditions, with a total of 22,869 scenarios covering all combinations of environment and vehicle factors in Table \ref{table:dimensions}).
For the \textit{driving dynamics}, we create a set of 100 driving cycles that capture a wide range of driving behaviors using the following five distinct strategies.
\begin{itemize}
    \item Random speed: A random walk generated by sampling random acceleration values from a uniform distribution with realistic acceleration range from -1.5 to 3 m/s$^2$. 
    The speed is truncated at 0 m/s to prevent a random acceleration that leads to negative speed values (Figure \ref{fig:traj}(a)).
    
    \item Sinusoidal speed: Periodic speed variations designed to enforce  acceleration and deceleration cycles with noise, mimicking stop-and-go traffic (Figure \ref{fig:traj}(b)). 

    \item Piecewise speed: A vehicle maintains a constant speed for a finite duration, followed by a transition period during which it reaches the next speed with constant acceleration. The set of constant speeds, their duration, and the length of the transition periods are parameterized and randomly set, allowing for variability in these three aspects while ensuring realistic acceleration values (Figure \ref{fig:traj}(c)). Such speed profiles are commonly used in human-compatible control applications \citep{mayuri-pwc}.
    
    \item Approaching intersections: The well-known Intelligent Driver Model (IDM) \citep{Treiber2000CongestedTS} is used to simulate a vehicle approaching a signalized intersection, where car-following, acceleration, deceleration, and idling behaviors are replicated (Figure \ref{fig:traj}(d)).

    \item Eco-driving: Dynamic eco-driving strategies are implemented to the vehicles that approach a signalized intersection (Figure \ref{fig:traj}(e)). A set of fuel and emission optimized driving cycles generated using reinforcement learning-based policy as outlined in \cite{jayawardana-eco}. The learned vehicle control policy reduces idling time at intersections by adjusting vehicle acceleration to synchronize its arrival with the green phases of the traffic signal. This results in a gliding behavior, minimizing the need for stopping and improving overall fuel and emission efficiency.

\end{itemize}
Each combination of the driving cycles, and environments and vehicles is processed with MOVES and NeuralMOVES, respectively, and total emissions are computed for comparison, rendering an extensive validation set of 22,869$\times$100=2,296,900 evaluations. 

\begin{figure}
    \centering
    \includegraphics[width=1\textwidth]{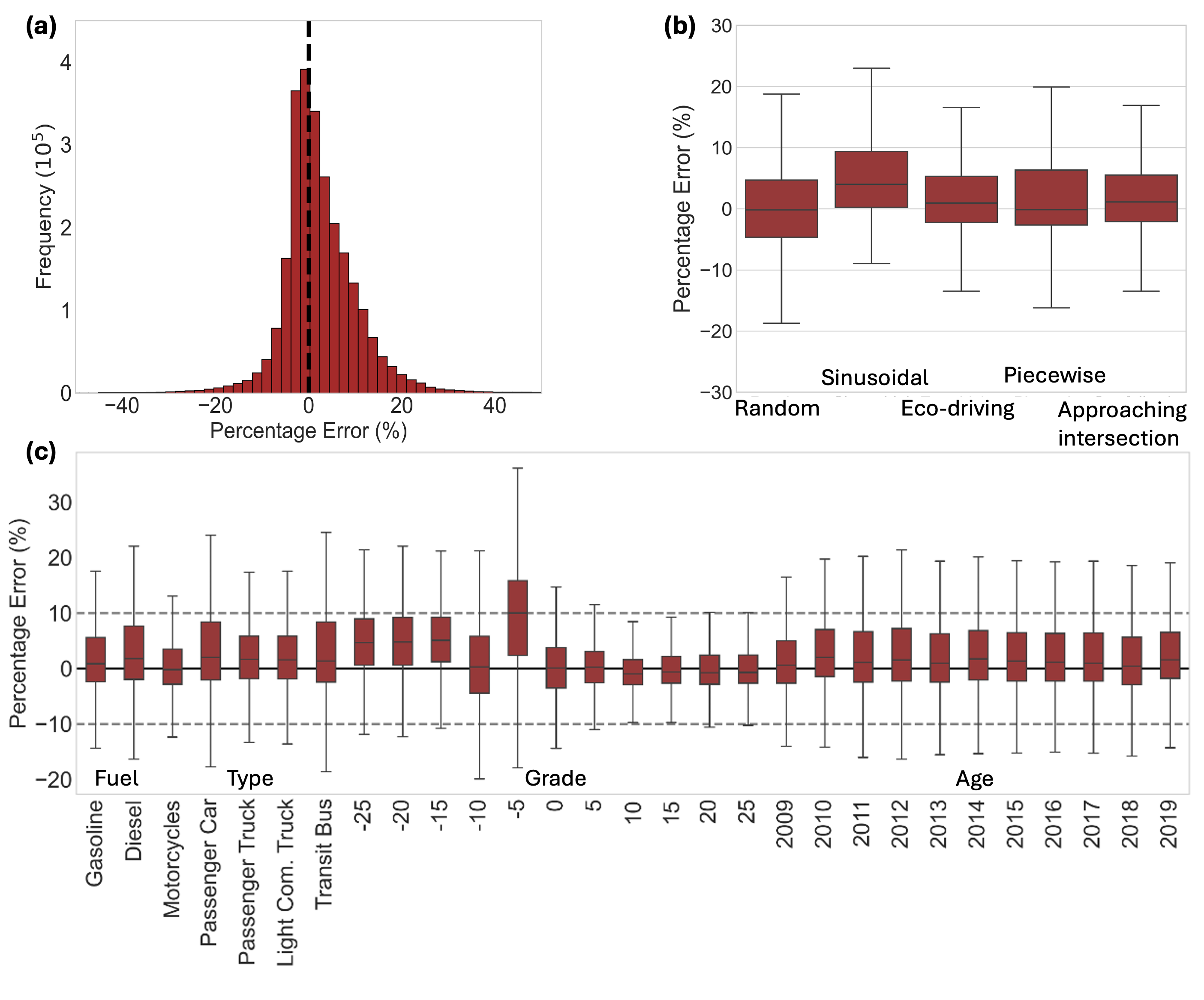}
    \caption{Emission estimation error of NeuralMOVES against MOVES under various scenarios. 
    (a) Percentage error distribution for all trajectories;
    (b) Percentage error distribution for the trajectories generated by various strategies;
    (c) Percentage error distribution for the trajectories generated under various environments and vehicle factors.}
    \label{fig:PE}
\end{figure}

The accuracy of NeuralMOVES is evaluated using MAPE across all 2,286,900 evaluations, yielding an overall MAPE of 6.013\%. 
Moreover, Figure~\ref{fig:PE} presents a detailed breakdown of the percentage error distribution across various dimensions. 
It is found that, although the maximum errors are about $\pm20\%$ (Figure~\ref{fig:PE}(a)), most of the errors centered at zero with most errors ranging between -5\% and 10\% under various environment and vehicle conditions (Figure~\ref{fig:PE}(c)).
The distribution exhibits slight positive skewness, indicating that NeuralMOVES is likely to overestimate rather than underestimate emissions (Figures~\ref{fig:PE}(a) and \ref{fig:PE}(c)).
Regarding the estimation under various scenarios, the error remains relatively consistent across the trajectories generated by distinct strategies (Figure~\ref{fig:PE}(b)).
In general, these results demonstrate that while NeuralMOVES introduces some error, it maintains high precision and consistent accuracy in replicating MOVES outputs across diverse scenarios. The observed biases and variability patterns highlight potential areas for refinement but confirm the surrogate model’s reliability as an efficient alternative for emissions estimation.

\section{Use case: Dynamic eco-driving}\label{sec:UseCase}

To showcase the practical benefits of NeuralMOVES and its effectiveness, we implement a Model Predictive Control (MPC)-based dynamic eco-driving application.
Dynamic eco-driving refers to optimizing driving behavior in real-time to minimize fuel consumption and emissions, adapting to traffic conditions and road characteristics \citep{Xia01012013,mintsis2020dynamic,YU2021103101}.
Dynamic eco-driving optimization is a typical traffic optimization problem that heavily involves the estimation of fuel consumption and/or emissions, as these variables are usually part of the objective function and need to be computed for each possible action at each time step of the optimization process.  \citep{HE2015106,ZHANG2023103607,LU2024104270,10693866}.

Dynamic eco-driving can be formulated as the following optimal control problem.
\begin{equation}
    \min_a \textbf{E} = \sum_{t=t_0}^{t_0+N} \left( w_e \mathbf{M}(a(t), v(t)) - w_v v(t) \right)
\end{equation}

subject to:
\begin{align}
x(t+1) &= x(t) + \frac{v(t)+v(t+1)}{2}\mathrm{d}t \label{equ:c1}\\
v(t+1) &= v(t) + a(t)\mathrm{d}t \label{equ:c2}\\
x(0) &= q_1 \label{equ:c3}\\
v(0) &= q_2 \label{equ:c4}\\
x(t) &\geq x(0) \quad \forall t \in [t_0, t_0+N] \label{equ:c5}\\
v_{\min} \leq v(t) &\leq v_{\max} \quad \forall t \in [t_0, t_0+N] \label{equ:c6}\\
a_{\min} \leq a(t) &\leq a_{\max} \quad \forall t \in [t_0, t_0+N] \label{equ:c7}\\
q_3 \leq x(N) &\leq q_4 \label{equ:c8} \\
v(N) &\geq q_5 \label{equ:c9}
\end{align}
where $\textbf{M}(\cdot)$ represents any emission estimation model; 
weights $w_e$ and $w_v$ balance emission minimization against speed maximization;
$v_{\min}$ and $v_{\max}$ are the lower and upper boundaries of vehicle speed;
$a_{\min}$ and $a_{\max}$ are the lower and upper boundaries of vehicle acceleration;
$q_1$, $q_2$, $q_3$, $q_4$ and $q_5$ are all coefficients.
The constraints define 
vehicle dynamics (Constraints \eqref{equ:c1} and \eqref{equ:c2}), 
initial conditions (Constraints \eqref{equ:c3} and \eqref{equ:c4}), 
and operational bounds (Constraints \eqref{equ:c5} $\sim$ \eqref{equ:c9}).

MPC iteratively solves the finite-horizon optimal control problem to determine acceleration sequences that minimize cumulative emissions while satisfying dynamic constraints. 
Specifically, the vehicle's initial state including position, velocity, and environmental conditions, is established. 
Then, at each timestep, MPC solves the optimization problem to determine the sequence of control actions \( \mathbf{a}_{0:T-1} \) that minimizes the cumulative cost over the horizon \( T \), where the cost function incorporates emission predictions from NeuralMOVES to consider environmental impact. Only the first control action \( a_0 \) is then applied to the vehicle, updating its state based on the control action and vehicle dynamics model. This process is repeated at each time step using the updated state information.

\begin{figure}
    \centering
    \subfigure[Car: grade]{\includegraphics[width=0.32\textwidth]{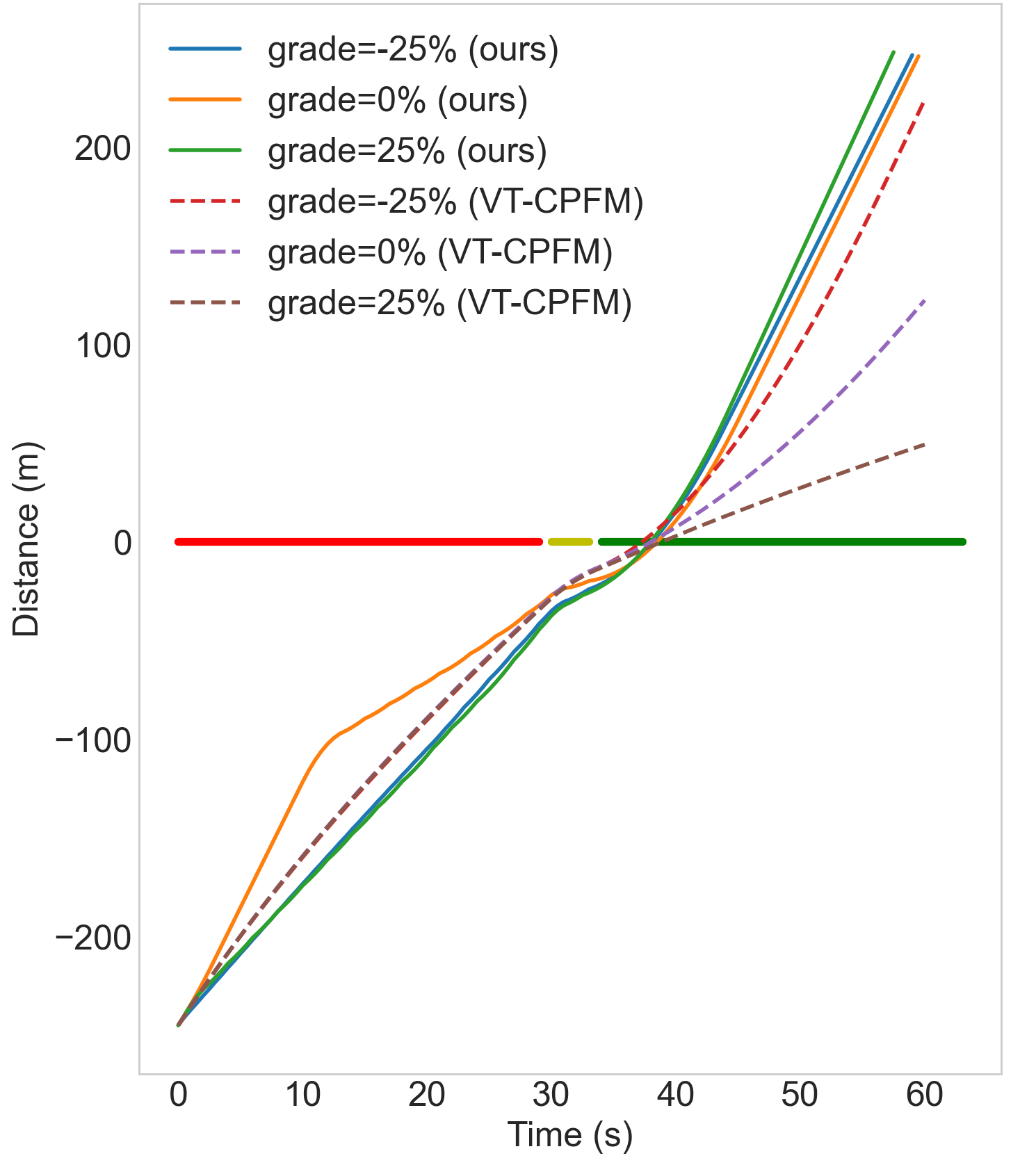}}
    \subfigure[Car: temperature]{\includegraphics[width=0.32\textwidth]{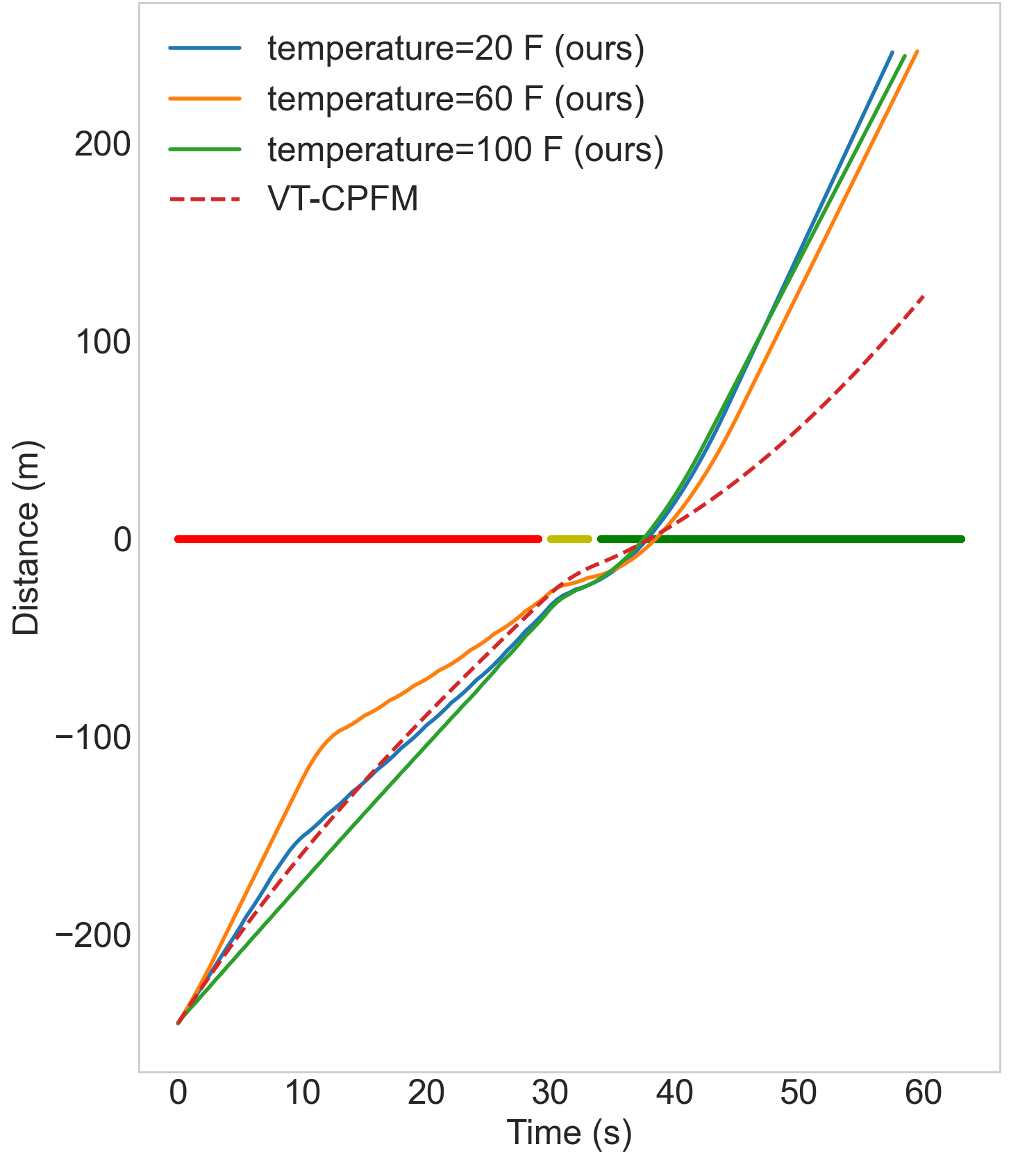}}
    \subfigure[Car: Humidity]{\includegraphics[width=0.32\textwidth]{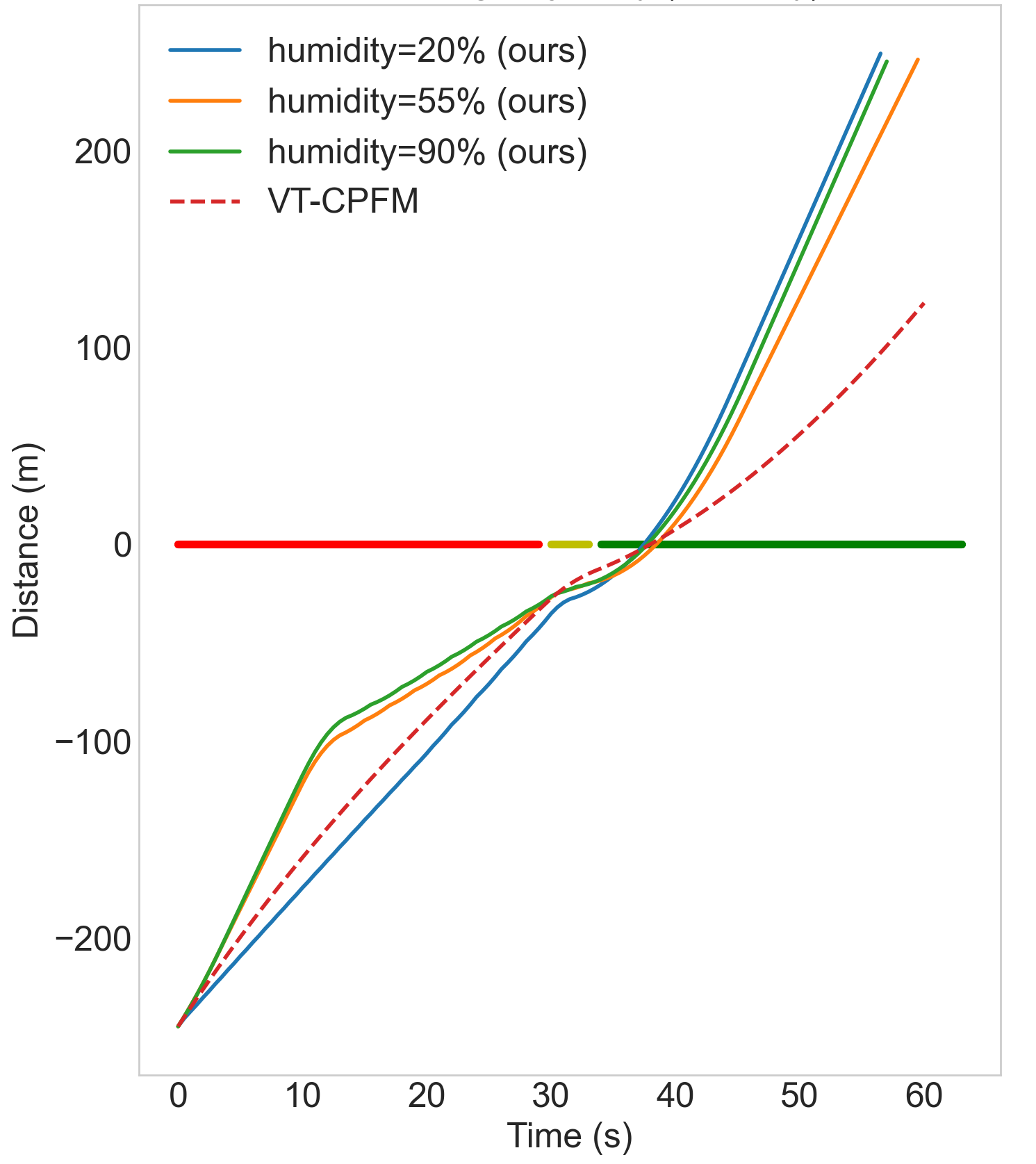}}
    \subfigure[Bus: grade]{\includegraphics[width=0.32\textwidth]{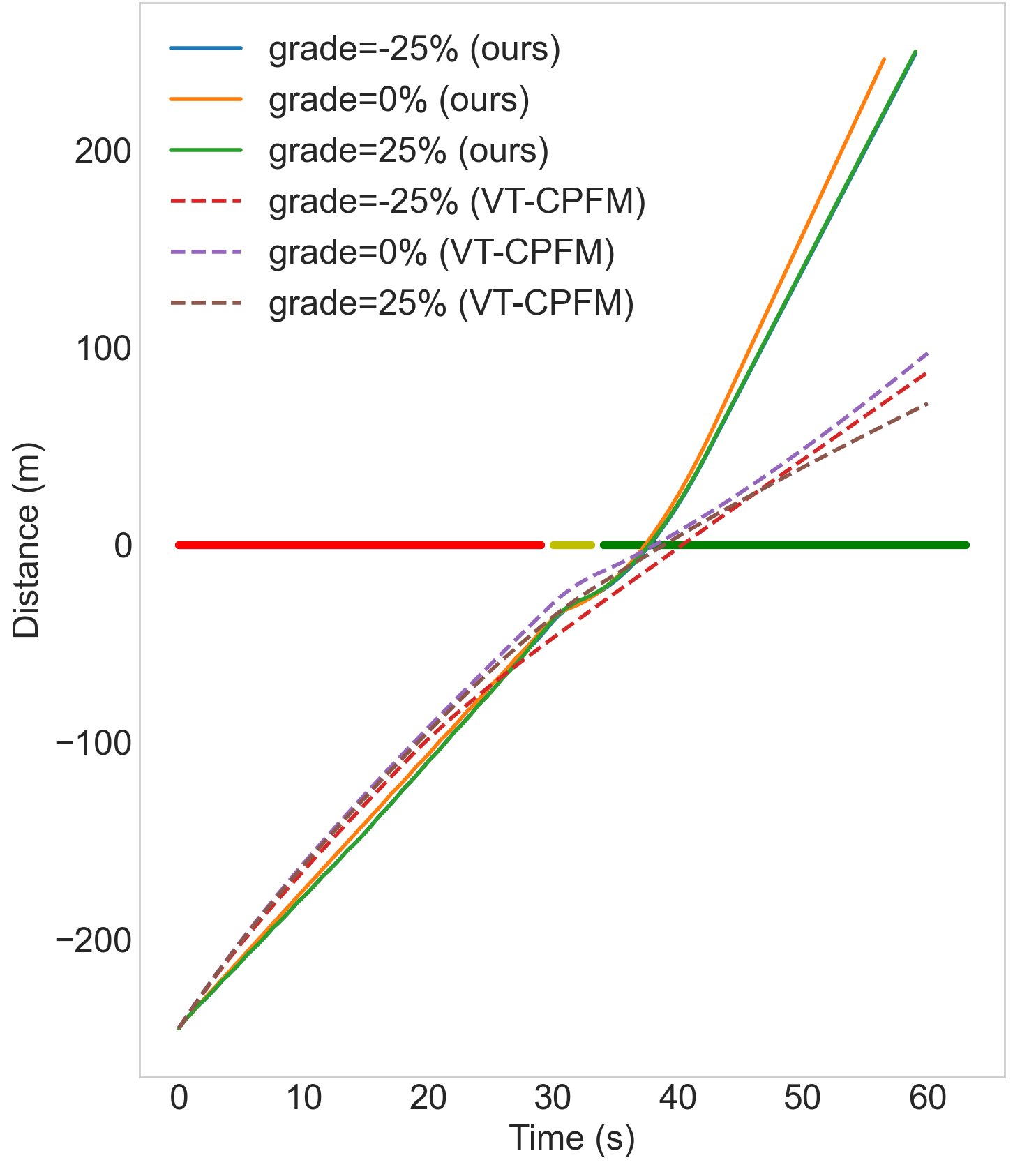}}
    \subfigure[Bus: temperature]{\includegraphics[width=0.32\textwidth]{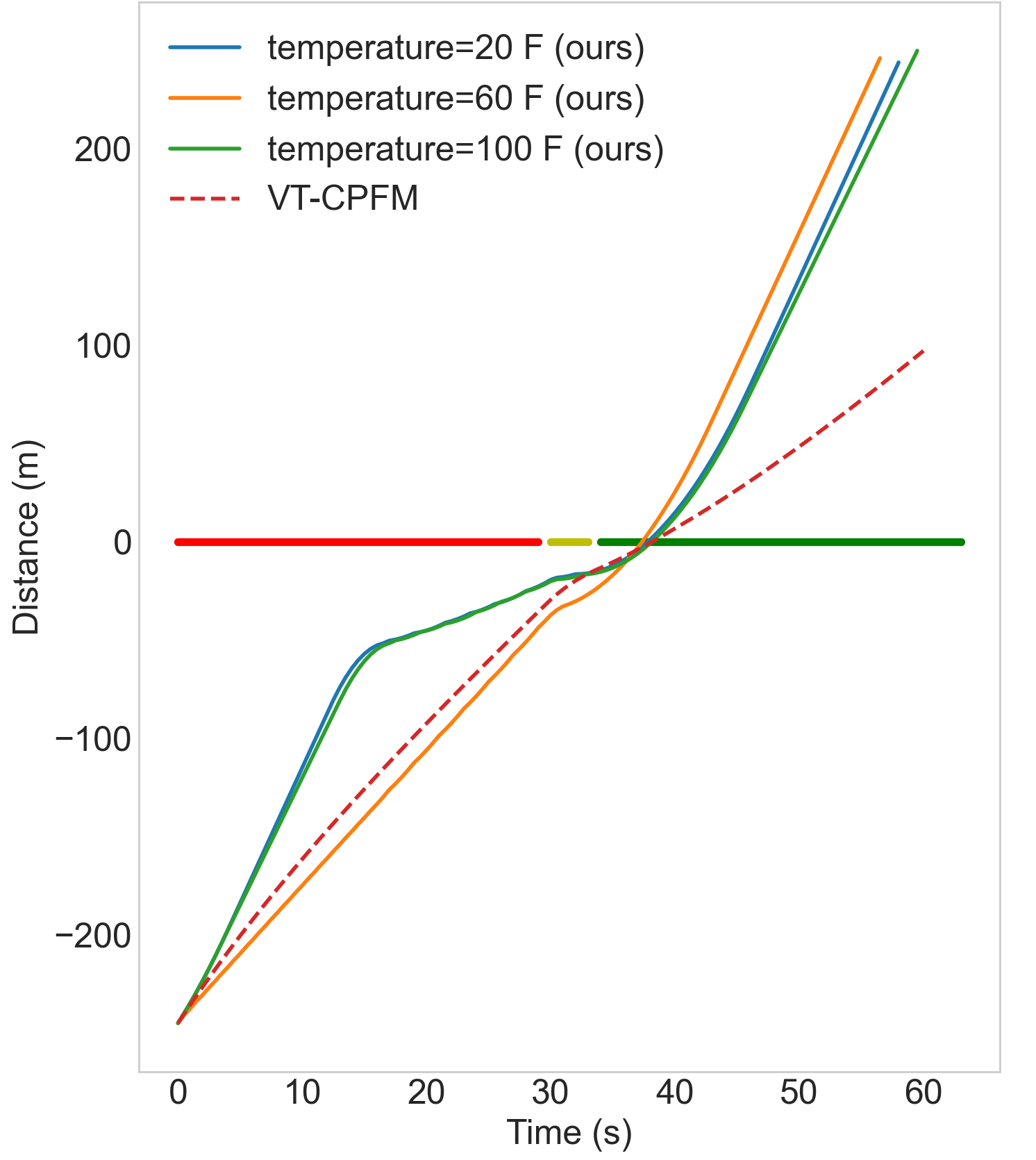}}
    \subfigure[Bus: Humidity]{\includegraphics[width=0.32\textwidth]{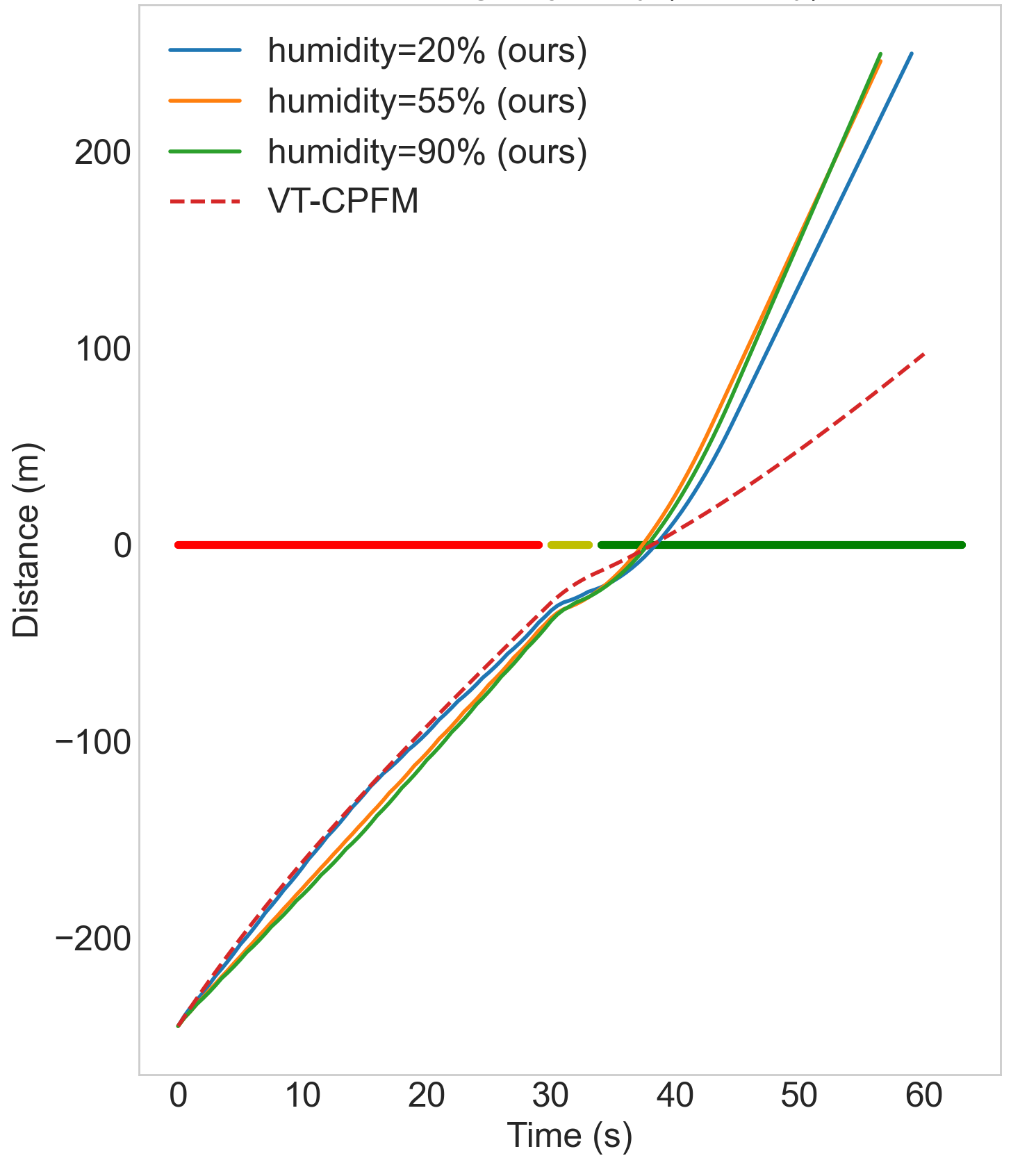}}
    \caption{Optimized trajectories for 2019 passenger car model (a)-(c) and 2019 bus model (d)-(f) given by NeuralMOVES and VT-CPFM. 
    To analyze the effect of each parameter individually, we vary one parameter at a time (grade, temperature, or humidity) while keeping the others constant at fixed values of 0 \%, 60 $^\circ$F, and 55 \% for grade, temperature, and humidity, respectively.}
    \label{fig:car_bus}
\end{figure}

Figures~\ref{fig:car_bus} present a series of optimized driving trajectories generated for a 2019 passenger car model and a 2019 bus model in the scenario of approaching a signalized intersection. 
These trajectories were calculated under three distinct scenarios: varying road grades, ambient temperatures, and humidity levels. 
In each scenario, the surrogate model efficiently adjusted to account for external factors, producing driving paths that minimize CO$_2$ emissions while adhering to travel time requirements. 
For instance, steeper road grades necessitate shifts in acceleration patterns, whereas temperature and humidity changes influenced engine performance and emissions profiles.

Our experimental results reveal that optimal trajectories vary significantly with environmental parameters, demonstrating that eco-driving strategies are indeed environment-dependent. This finding underscores the importance of using emission models that can accurately capture these environmental dependencies. Furthermore, the trajectories generated using NeuralMOVES differ notably from those produced by simpler emission models, such as VT-CPFM, suggesting that previous eco-driving applications might have yielded suboptimal results by not incorporating comprehensive emission modeling.

At last, we emphasize that the following two key properties of NeuralMOVES make it compatible and succuessful with this optimization technique and other control and optimization applications.
\begin{itemize}
    \item NeuralMOVES is differentiable, a crucial property for gradient-based optimization methods. In contrast, MOVES and other dataset-based emission estimation models are not differentiable, making them unsuitable for integration into the optimization frameworks. As a Neural Network-based surrogate model, NeuralMOVES enables the computation of gradients with respect to control inputs, thereby facilitating the optimization process.

    \item NeuralMOVES has a computation time on the order of milliseconds for instantaneous emission estimates (i.e., one forward pass of the Neural Network). This rapid emission estimation enables real-time application within an optimization method. 
    Such computational efficiency is critical, as many optimization and control methods require frequent evaluations of the cost function to determine optimal control actions at each timestep.
\end{itemize}

\section{Conclusions and discussion}\label{sec:Conclusion}

MOVES, as one of the most well-established emission estimation models, serves as the official and state-of-the-art emission estimation model in the U.S..
While a powerful and authoritative tool, MOVES presents significant challenges due to its complexity, data requirements, and computational demands, which limit its accessibility and applicability.
To address these limitations, this paper extracts vehicle running CO$_2$ emissions through carefully designed data extraction and scenario generation.
The reverse engineering approach ultimately generates a dataset of more than 9.89 GB with 109,367,240 data points, representing MOVES.
Furthermore, we proposed a surrogate learning-based method to approximate the reverse engineered MOVES dataset and develop a computationally efficient lightweight model, namely, NeuralMOVES. Importantly, the 2.4 MB NeuralMOVES achieves only a 6.013\% MAPE compared to MOVES across diverse scenarios, while offering a differentiable emission estimation model with a computing time in the order of milliseconds.

A dynamic eco-driving problem is solved as a use case, demonstrating the capability and flexibility of the proposed NeuralMOVES, which is well-suited (but not limited) to computationally intensive real-time and microscopic applications.

The proposed NeuralMOVES represents a significant advancement in emission modeling for the transportation sector, contributing to both macro- and micro-scale analyses.
Its scalability and adaptability make it a valuable resource for both researchers and practitioners.
At last, it is important to note that the current NeuralMOVES only accounts for CO$_2$ emissions from vehicle running operation. Future work could include extracting air pollutants, such as NO$_x$ and PM$_{2.5}$, and emissions from other driving states, such as starting emissions. Moreover, the accuracy of the model could be further improved.

\bibliographystyle{apalike} 
\bibliography{reference}
\end{document}